\documentclass[final]{cvpr}

\usepackage{times}
\usepackage{epsfig}
\usepackage{graphicx}
\usepackage{amsmath}
\usepackage{amssymb}
\usepackage{multirow}
\usepackage{colortbl}
\usepackage{color}

\usepackage[pagebackref=true,breaklinks=true,colorlinks,bookmarks=false]{hyperref}

\def\cvprPaperID{****} 
\def\confYear{CVPR 2021}

\begin{document}

\title{Revisiting Classification Perspective on Scene Text Recognition}

\author{
 Hongxiang Cai\thanks{Equal contribution.}, Jun Sun\footnotemark[1] , Yichao Xiong\thanks{Corresponding author.} \\
  Media Intelligence Technology Co.,Ltd\\
  \texttt{\small \{hongxiang.cai, jun.sun, yichao.xiong\}@media-smart.cn} 
}

\maketitle

\begin{figure*}[ht]
\centering
  \includegraphics[width=1.0\textwidth]{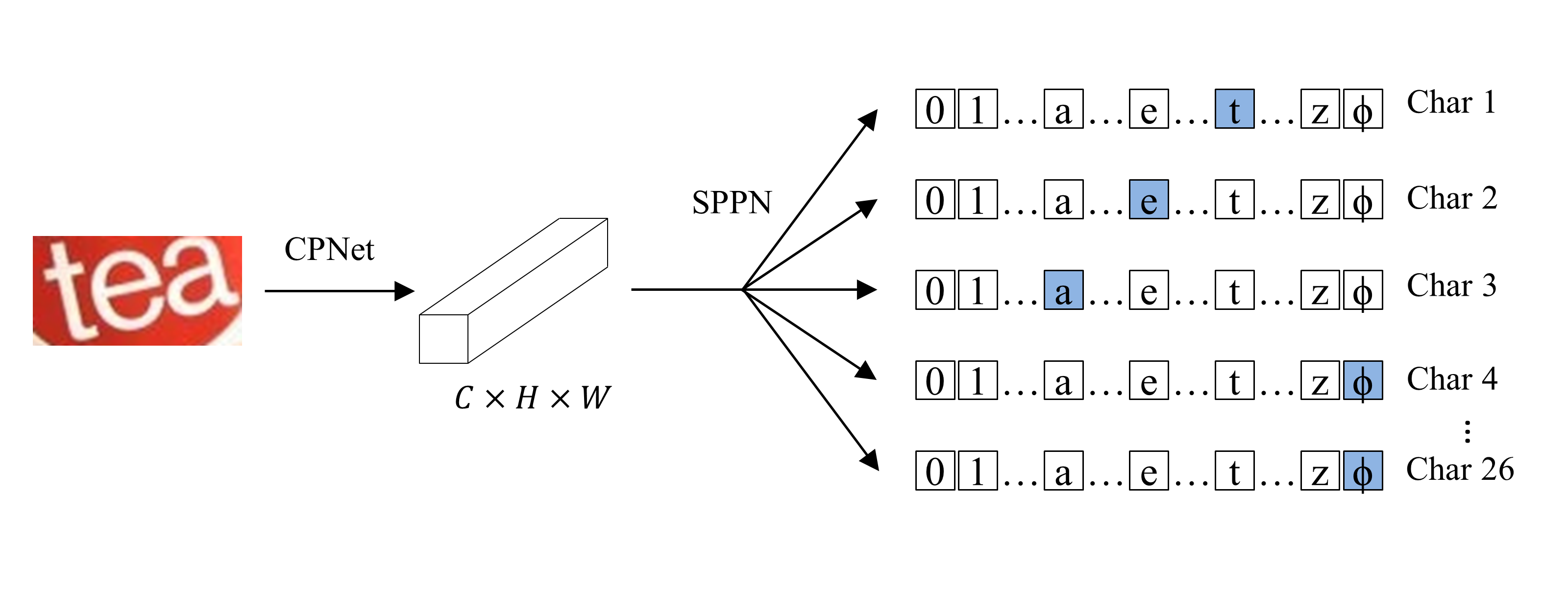}
\centering
\caption{The model architecture of CSTR. $\phi$ indicates the end token.}
\label{fig.CSTR_model}
\end{figure*}

\begin{abstract}
The prevalent perspectives of scene text recognition are from sequence to sequence (seq2seq) and segmentation. Nevertheless, the former is composed of many components which makes implementation and deployment complicated, while the latter requires character level annotations that is expensive.
In this paper, we revisit classification perspective that models scene text recognition as an image classification problem. Classification perspective has a simple pipeline and only needs word level annotations. We revive classification perspective by devising a scene text recognition model named as CSTR, which performs as well as methods from other perspectives.
The CSTR model consists of CPNet (classification perspective network) and SPPN (separated conv with global average pooling prediction network). 
CSTR is as simple as image classification model like ResNet \cite{he2016deep} which makes it easy to implement and deploy.
We demonstrate the effectiveness of the classification perspective on scene text recognition with extensive experiments. Futhermore, CSTR achieves nearly state-of-the-art performance on six public benchmarks including regular text, irregular text. The code will be available at \url{https://github.com/Media-Smart/vedastr}.
\end{abstract}

\section{Introduction}
Text is one of the most important carrier of information and knowledge, it can be found almost everywhere in our life such as books, newspapers, road signs, billboards, etc. The objective of scene text recognition (STR) is to translate a cropped text instance image into a target string sequence, which is very useful in a range of applications, such as ID card scan, image-based machine translation, industrial automation, self-driving, to name a few. 

With the development of deep learning, a lot of neural network based methods significantly boosted the performance of STR in which two categories can be divided: segmentation-based methods \cite{liao2019scene} and seq2seq-based methods.
Seq2seq-based methods can be roughly classified into CTC-based \cite{graves2006connectionist} methods \cite{shi2016end, borisyuk2018rosetta, du2020pp}, attention-based \cite{bahdanau2014neural} methods \cite{shi2016robust, shi2018aster, li2019show, zhan2019esir}, transformer-based \cite{vaswani2017attention} methods \cite{sheng2019nrtr, yang2019simple, lu2019master, lee2020recognizing, bartz2019kiss, yu2020towards}.

Segmentation-based \cite{liao2019scene} methods usually include two steps: character segmentation,
and character recognition. Methods in this category attempt to locate the position of each character on the input text instance image, apply a character classifier to recognize each character, and group characters into text line to obtain the final recognition result.

Seq2seq-based methods recognize the text line as a whole and focus on mapping the entire text instance image into a target string sequence directly by an encoder-decoder framework, thus avoiding character segmentation. This approach contains two stages: image encoding and sequence decoding. In image encoding phase, the input text instance image will be encoded into a feature sequence by Convolutional Neural Network (CNN) \cite{lecun1998gradient}, Recurrent Neural Network (RNN), transfomer \cite{vaswani2017attention} encoder or the combinations. The feature sequence then will be decoded by CTC \cite{graves2006connectionist}, attention based RNN or transformer \cite{vaswani2017attention} decoder into text line.

Segmentation-based methods \cite{liao2019scene} are simple, but they require expensive character level annotations. Many systems based on seq2seq have achieved state-of-the-art performance, however seq2seq-based methods are complex and have a long pipeline.

In addition to above prevalent two perspectives, there are also a minority in the literature which model STR as an image classification problem which can be named as classification-based method. CHAR \cite{jaderberg14c} assumes that there is a maximum number $k$ of characters per word  that can be recognized.
The body of the CHAR model consists of four convolutional layers and two fully connected layers followed by $k$ independent multi-class classification heads, each of which predicts the character at each position. An end token $\phi$ is introduced  to handle the problem that words have variable length which is unknown at test time.
The CHAR model is easy to train like an image classification model, but its accuracy is very low.

Classification-based methods \cite{jaderberg14c} are really simple which makes it very attractive. However, CHAR performs badly which leads to few following works from classification perspective \cite{borisyuk2018rosetta}. In this paper, we revisit classification-based methods and demonstrate that this perspective can achieve comparable performance with seq2seq-based methods and segmentation-based methods through our carefully designed CSTR.

Specifically, we devise a dedicated prediction network, SPPN (separated conv with global average pooling prediction network), which incorporates global semantic information to implicitly encode character position of text sequence. This design match the mechanism behind classification perspective that $i$-th classification head predicts $i$-th character of the word sequence in the input image, where the classification head should know the $i$ characters from the left. SPPN is the key component that makes classification-based methods work better than CTC-based methods with low back-propagating computation burden under the same number of parameters.

Furthermore, we design a new backbone network, CPNet (classification perspective network), which not only improves the network's ability to focus on important features and suppress unnecessary one but also has larger valid receptive field. This powerful backbone network, which coordinates classification perspective, brings a large improvement for CSTR.

CSTR achieves nearly state-of-the-art performance on a variety of standard scene text recognition benchmarks such as ICDAR03, ICDAR13, ICDAR15, IIIT5k, SVT, SVTP.

Our main contributions are highlighted as follows:
\begin{itemize}

\item We demonstrate that cross entropy (CE) loss perform better than CTC loss when prediction network is devised properly by thorough experiments.

\item We propose a novel method named CSTR which is the first classification-based method on STR that works as well as segmentation-based methods and seq2seq-based methods.

\item CSTR achieves nearly state-of-the-art accuracy on benchmarks for scene text recognition.

\item The proposed model is as simple as image classification model and effective.
\end{itemize}

\section{Related Work}
Over  the past few years, the research of scene text recognition has made significant progress with the development of deep learning. In this section, we will review some recent scene text recognition methods, and group the prevalent of them into two categories: seq2seq-based and segmentation-based. Besides, the classification-based methods which are nearly ignored by the community and some popular modules will also be involved.

\textbf{Seq2seq-based methods} recognize the text line as a whole and focus on mapping the entire text instance image into a target string sequence directly by a encoder-decoder framework. This approach contains two stages: image encoding and sequence decoding. In image encoding phase, the input text instance image will be encoded into a feature sequence by CNN, RNN, transfomer \cite{vaswani2017attention} encoder or their combinations. The feature sequence then will be decoded by CTC \cite{graves2006connectionist}, attention \cite{bahdanau2014neural} based RNN or transformer \cite{vaswani2017attention} decoder into text line. 

CRNN \cite{shi2016end}, one of the first seq2seq-based scene text recognition methods, uses CNN to extract features which are encoded into a feature sequence by RNN, then CTC  \cite{graves2006connectionist} is introduced to decode the feature sequence into text line.

The idea of using a recurrent neural network to predict
a character sequence has since been extended by various
methods that incorporate an attention mechanism \cite{bahdanau2014neural, chorowski2015attention} into the
text sequence decoding \cite{shi2016robust, shi2018aster, li2019show, zhan2019esir}. This line of work extracts visual features by a convolutional neural network which is encoded into a feature sequence by a recurrent neural network, then another recurrent neural network equipped with attention mechanism is adopted to decode the feature sequence into text line. 

Recently, the structure of transformer \cite{lin2017structured} has been proposed to capture global dependencies. Transformer has been proved to be effective in many tasks of computer vision \cite{wang2018non, carion2020end} and natural language processing \cite{vaswani2017attention, devlin2018bert, brown2020language}. A lot of work \cite{sheng2019nrtr, yang2019simple, lu2019master, lee2020recognizing, bartz2019kiss, yu2020towards} incorporates transformer into scene text recognition and achieves good results. They use transformer encoder to encode visual features into a feature sequence which is then decoded into text line by transformer decoder. 
Seq2seq-based methods are the most popular in scene text recognition community, however they are complex.

\textbf{Spatial transformer} \cite{jaderberg2015spatial} not only selects regions of an image that are most relevant, but also transforms those regions to a canonical, expected pose (horizontally aligned characters of uniform heights and widths) to simplify recognition in the following layers.

The spatial transformer is split into three parts: localisation network, grid generator and sampler. First, a localisation network takes the input image, and outputs the parameters of the spatial transformation that should be applied to the input image. Then, the predicted transformation parameters are used to create a sampling grid, a set of points where the input image should be sampled to produce the transformed output image, which is done by the grid generator. Finally, the input image and the sampling grid are taken as inputs to the sampler, producing the output rectified image sampled from the input image at the grid points.

ASTER \cite{shi2018aster} adds a spatial transformer before the scene text recognition model to eliminate the negative effects of perspective distortion and distribution curvature which boosts performance on irregular texts. Since the proposal of spatial transformer in \cite{liu2016star}, spatial transformer has been a default module  on scene text recognition \cite{shi2016robust, gao2018recurrent, shi2018aster, yang2019symmetry, zhan2019esir, hu2020gtc, litman2020scatter, qiao2020seed}.

\textbf{Segmentation-based methods} attempt to locate the position of each character on the input text instance image, apply a character classifier to recognize each character, and group characters into text line to obtain the final recognition results. CA-FCN \cite{liao2019scene} exploits extra character level annotations in addition to the word level annotations to supervise the training process which can overcome the disadvantage of seq2seq-based methods that missing or superfluous characters will cause misalignment between the ground truth strings and the predicted sequences.

While segmentation-based methods are simple, they require character level annotations which are expensive.

\textbf{Classification-based methods} model scene text recognition as an image classification problem which is simpler than seq2seq-based methods. Furthermore, classification-based methods don't require character level annotations like segmentation-based methods, which are hard to obtain.

CHAR \cite{jaderberg14c} assumes that there is a maximum number $k$ of characters per word  that can be recognized.
The body of the CHAR model consists of four convolutional layers and two fully connected layers followed by $k$ independent multi-class classification heads, each of which predicts the character at each position. An end token $\phi$ is introduced to handle the problem that words have variable length which is unknown at test time.
The CHAR model is easy to train like an image classification model, but it has a very low accuracy \cite{borisyuk2018rosetta}.

In this paper, we delve into classification-based methods and devise CSTR which demonstrates that classification-based methods can perform as well as seq2seq-based methods and segmentation-based methods with simpler architecture and no need for character level annotations. Moreover, CSTR abandons the spatial transformer which simplifies scene text recognition pipeline further more.

\section{Classification Perspective on STR}

Our aim is to design a simple and effective scene text recognition method which can be easily implemented and deployed like CTC which is widely used and practical in production \cite{borisyuk2018rosetta, du2020pp}.

CTC is a way to get around not knowing the alignment between the input and the output which is an intermediate process. We believe that the intermediate process is not necessary which is elaborated in Section \ref{section.from_CTC_loss_to_CE_loss}. We model the scene text recognition as image classification problem to avoid the alignment between input and output as is shown in Figure \ref{fig.CSTR_model}.

Borisyuk \textit{et al.} \cite{borisyuk2018rosetta} conducted experiments to compare CHAR \cite{jaderberg14c} and CTC which shows that CHAR is inferior to CTC. We argue that CHAR's bad performance doesn't imply that classification-based methods can not work well. We propose CSTR, a classification-based method, which achieves comparable performance with seq2seq-based methods and segmentation-based methods.

\section{CSTR}

As is shown in Figure \ref{fig.CSTR_model}, the network architecture of CSTR is as simple as image classification models. It consists of two main parts:
backbone network and prediction network.
Specifically, the backbone network extracts features from an input image and the prediction network uses the features to predict the character at each position.

\subsection{Backbone Network}
\label{section.bacbone_network}
We use ResNet proposed by Cheng \textit{et al.} \cite{cheng2017focusing} as our basic backbone network. For each residual block, we use the projection shortcut (using $1 \times 1$ convolutions) when the dimensions between input and output are different, and use the identity shortcut when they are the same.
Due to different modeling method, we have to redesign the backbone network from classification perspective which leads to the birth of CPNet. The detailed architecture of CPNet is shown in Table \ref{table.backbone} without FPN.

\textbf{Depth, Width and Resolution.}
From the point of view of network depth, we deepen stage-3 and stage-4 and insert additional residual blocks in stage-5. With respect to network width, we widen it by 1.5 times.
In order to better exploit the model capacity, input image resolution is enlarged from $32\times128$ to $48\times192$ as described in \cite{tan2019efficientnet}.

\begin{figure}[ht]
\centering
  \includegraphics[width=0.45\textwidth]{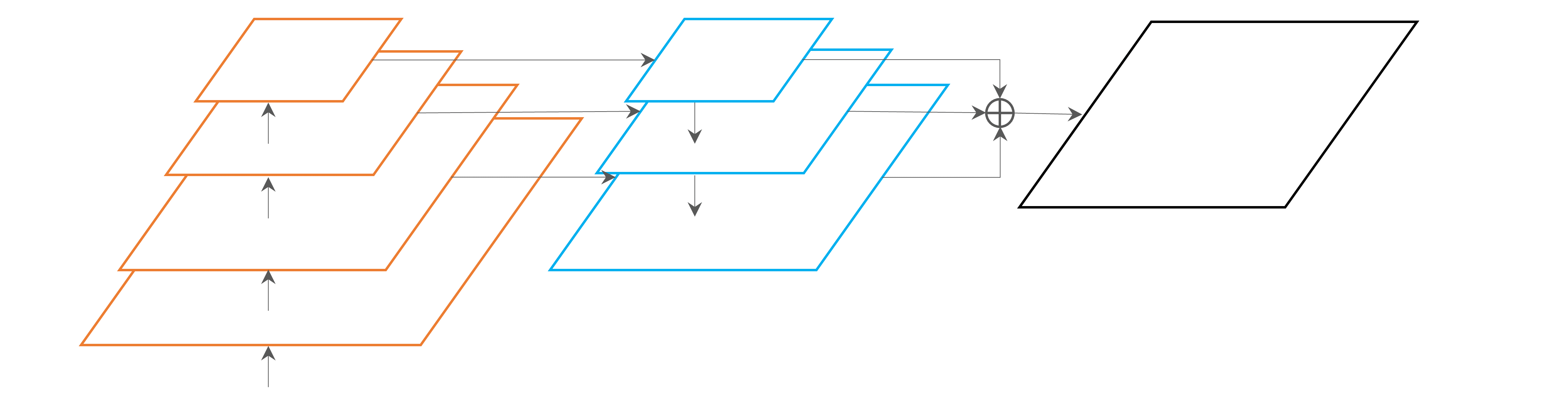}
\centering
\caption{The model architecture of FPN.}
\label{fig.FPN_model}
\end{figure}

\textbf{Feature Pyramid Network.}
Following Yu \textit{et al.} \cite{yu2020towards}, we also use FPN \cite{lin2017feature} to aggregate hierarchical feature maps from stage-3, stage-4 and stage-5 which is illustrated in Figure \ref{fig.FPN_model}. Thus the output feature map width and height are 1/4 of the input image, and the channel number is 512.

\textbf{CBAM.}
Attention \cite{woo2018cbam} has been proved effective in classification. Since our proposed method is based on classification perspective, we insert attention modules \cite{woo2018cbam} in all residual blocks in order to focus on important features and suppress unnecessary ones.

\textbf{Semantic-Aware Downsampling Module.}
In addition to the above classic modules, we propose the semantic-aware downsampling module (SADM) to reduce discriminative information loss in feature downsampling procedure.

\begin{figure}[ht]
\centering
  \includegraphics[width=0.8\linewidth]{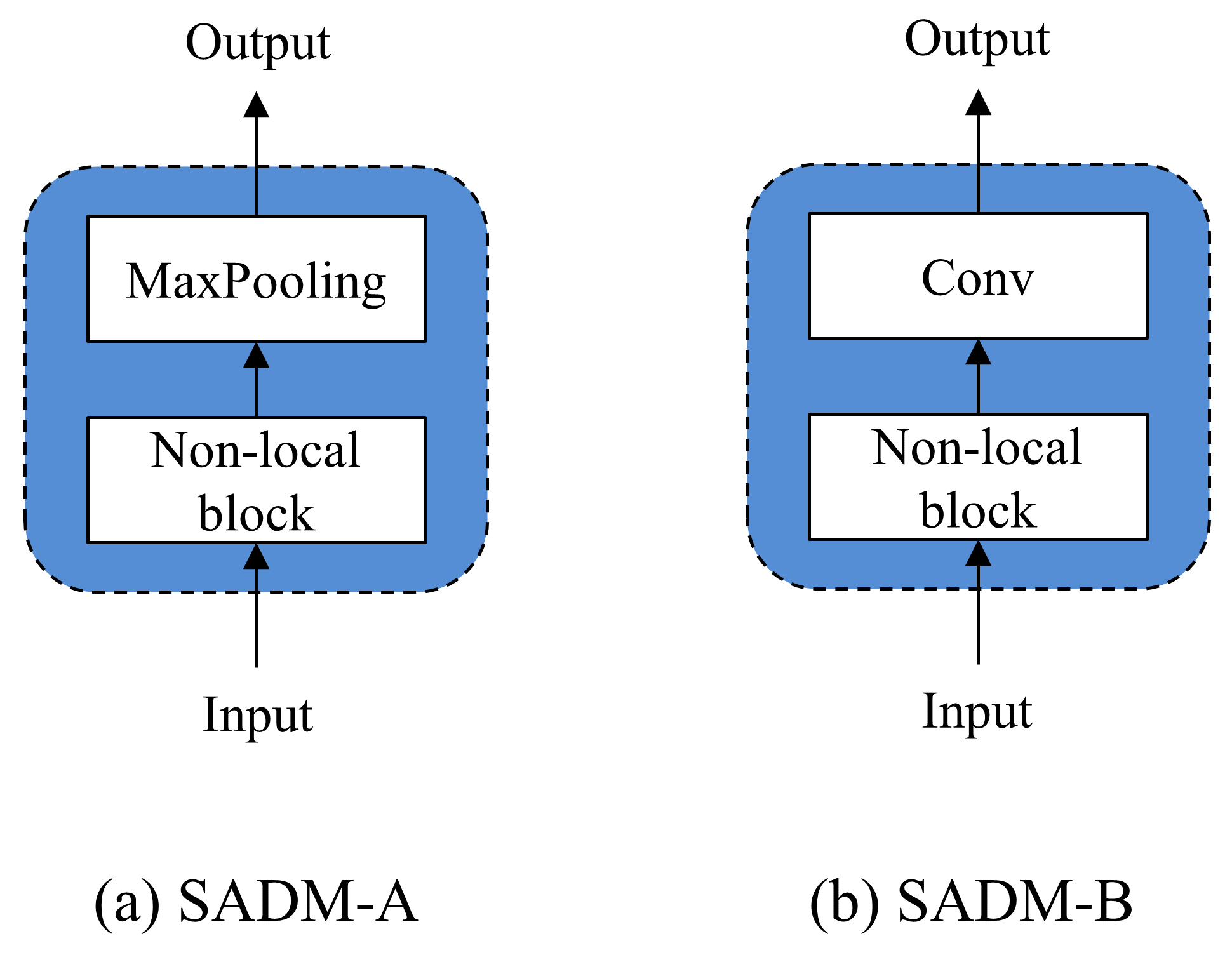}
\centering
\caption{The model architecture of SADM.}
\label{fig.SADM_model}
\end{figure}

We equip traditional downsampling module with non-local \cite{wang2018non} unit which can effectively capture global spatial interaction between features to make the network aware of which features should be discarded and what should be preserved when downsampling. Figure \ref{fig.SADM_model} depicts SADM details.

\begin{table}[ht]
\caption{CPNet body architecture. Residual blocks are highlighted with gray background.}
\label{table.backbone}
\centering
\resizebox{0.45\textwidth}{!}{
\begin{tabular}{|c|c|}
\hline
Stage name                & Type / Stride : Filter Shape                                      \\ \hline
                          & $Conv / s1: 1 \ \times 3 \ \times \ 3 \times \ 48$     \\ \cline{2-2} 
\multirow{-2}{*}{Stage 1} & $Conv / s1: 48 \ \times 3 \ \times \ 3 \times \ 96$           \\ \hline
                          & $Pool / s2: 2 \ \times \ 2$  \\ \cline{2-2} 
                          & \cellcolor[gray]{.95} $\left[
                                    \begin{array}{c}
						                Conv/s1: 96 \ \times \ 3 \ \times \ 3 \ \times \ 192\\
						                Conv/s1: 192 \ \times \ 3 \ \times \ 3 \ \times \ 192
						            \end{array}
						  \right] \times 1$ \\ \cline{2-2} 
\multirow{-3}{*}{Stage 2} & $Conv / s1: 192 \ \times 3 \ \times \ 3 \ \times \ 192$     
                                            \\ \hline
                          & $SADM$-$A$                                                \\ \cline{2-2}
                          & \cellcolor[gray]{.95} $\left[
                                \begin{array}{c}
						            Conv / s1: 192 \ \times \ 3 \ \times \ 3 \ \times \ 384\\
						            Conv / s1: 384 \ \times \ 3 \ \times \ 3 \ \times \ 384\\
						        \end{array}
						   \right] \times 4$    \\ \cline{2-2} 
\multirow{-3}{*}{Stage 3} & $Conv / s1: 384 \ \times \ 3 \ \times \ 3 \ \times \ 384$  \\ \hline
                          & $SADM$-$A$                                               \\ \cline{2-2}
                          & \cellcolor[gray]{.95} $\left[ 
                            \begin{array}{c}
						        Conv / s1: 384 \ \times \ 3 \ \times \ 3 \ \times \ 768\\
						        Conv / s1: 768 \ \times \ 3 \ \times \ 3 \ \times \ 768\\
						    \end{array}
						  \right] \times 7$                        \\ \cline{2-2} 
                          & $Conv / s1: 768 \ \times \ 3 \ \times \ 3 \ \times \ 768$ \\ \cline{2-2} 
\multirow{-4}{*}{Stage 4} & \cellcolor[gray]{.95} $\left[ 
                            \begin{array}{c}
						        Conv / s1: 768 \ \times \ 3 \ \times \ 3 \ \times 768\\
						        Conv / s1: 768 \ \times \ 3 \ \times \ 3 \ \times 768\\
						    \end{array}
						  \right] \times 5$                        \\ \hline
                          & $SADM$-$B$                                                \\ \cline{2-2} 
                          & \cellcolor[gray]{.95}$\left[ 
                          \begin{array}{c}
						  Conv / s1: 768 \ \times \ 3 \ \times \ 3 \ \times \ 768\\
						  Conv / s1: 768 \ \times \ 3 \ \times \ 3 \  \times \ 768
						  \end{array}
						  \right] \times 3$                        \\ \cline{2-2} 
\multirow{-4}{*}{Stage 5} & $Conv / s1: 768 \ \times 2 \ \times \ 2 \ \times \ 768$  \\ \hline
\end{tabular}
}
\end{table}







\subsection{Prediction Network}
\begin{figure*}[ht]
\centering
  \includegraphics[width=1.0\textwidth]{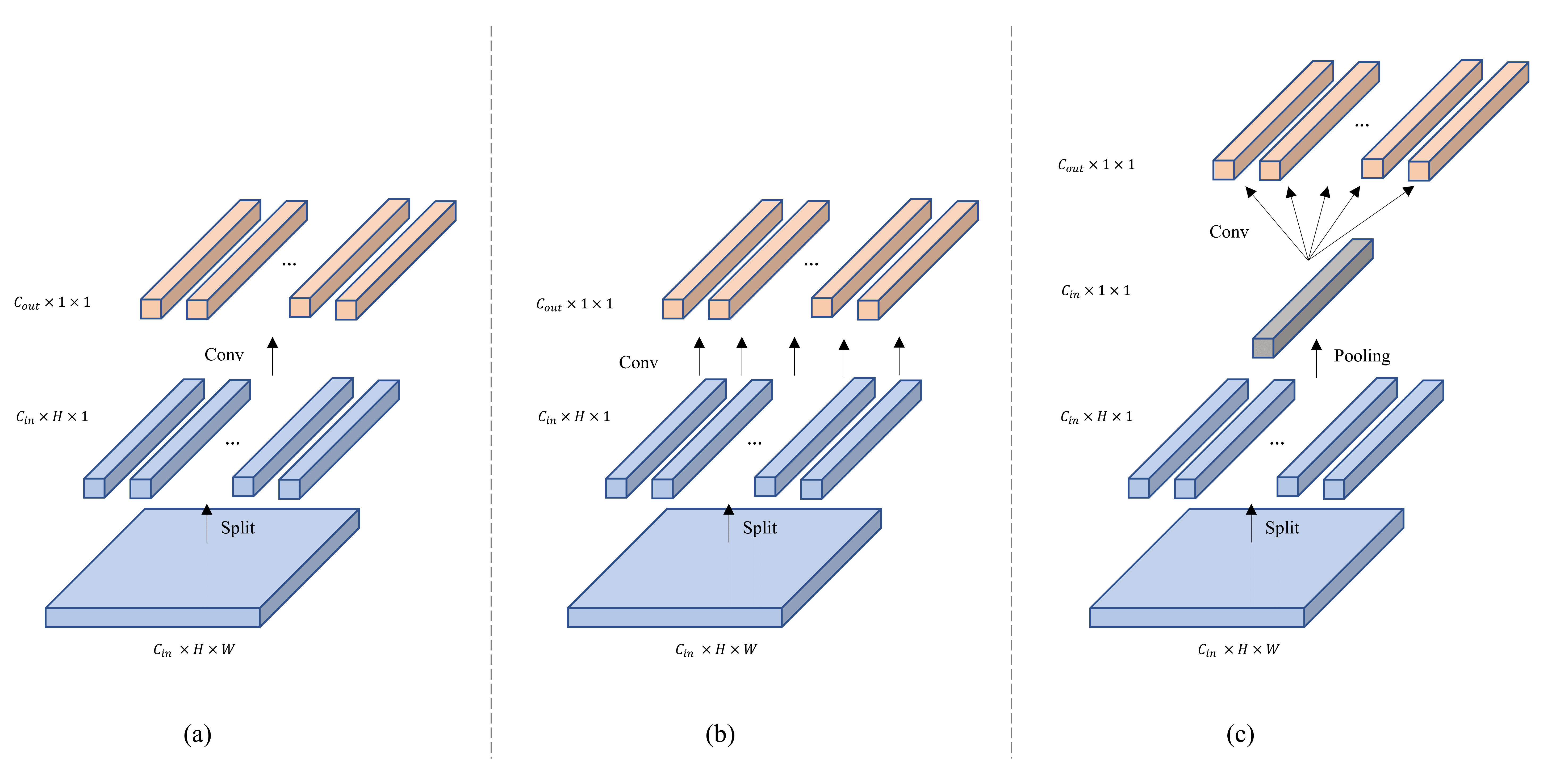}
\centering
\caption{The architecture of the prediction network. (a) SHPN. (b) SEPN. (c) SPPN. $C_{in}$ represents the input channels, $C_{out}$ represents the output channel, $H$ represents the height of the feature map, $W$ represents the width of the feature map.}
\label{fig.CTC_vs_CE}
\end{figure*}

A prediction network in CSTR is responsible for predicting the text line from output feature of backbone network. There are many choices of prediction network design. We survey the current popular methods in STR and classification, and design three types of prediction network which are shown in Figure \ref{fig.CTC_vs_CE}.

\textbf{Shared conv prediction network} (SHPN) uses a convolutional layer followed with backbone network which is shown in Figure \ref{fig.CTC_vs_CE}(a). If the dimension of the feature from backbone network is $C_{in} \times H \times W$, then the maximum length of prediction word is $W$ in this manner. This prediction network is widely used in CTC based architecture\cite{baek2019wrong, borisyuk2018rosetta}.

\textbf{Separated conv prediction network} (SEPN) uses separated convolutional layers in parallel after backbone network and we display it in Figure \ref{fig.CTC_vs_CE}(b). If the dimension of the feature from backbone network is $C_{in} \times H \times W$, the number of convolutional layers is $W$. It has the same property like SHPN that it can only predict $W$ length words. Compared with SHPN, SEPN has more parameters.

\textbf{Separated conv with global average pooling prediction network} (SPPN) utilizes a global average pooling layer after backbone network followed by separated convolutional layers. The potential design philosophy of SPPN is to introduce global semantic information to the next stage which has been proved effective by \cite{yu2020towards}. The architecture of SPPN is illustrated in Figure \ref{fig.CTC_vs_CE}(c).

We implement these three types of prediction network in CSTR respectively. Compared these three types of prediction network, SHPN and SEPN encode the character with positional information explicitly by splitting the input feature into $W$ features of which dimension is $C_{in} \times H \times 1$. Due to we modeling STR from classification perspective, each split feature should contain global semantic information to know the character order in the text line.
Nevertheless, restricted by the receptive field and the various distributions of character in image, SHPN and SEPN cannot make sure that the features satisfy the above requirements. However, with global average pooling we can incorporate global semantic information to the feature which implicitly encodes character position of text sequence.

It's worth noting that we don't adopt the prediction network used in CHAR\cite{bahdanau2014neural}, which used several fully connected layers in prediction network. The fully connected layers will result in lots of parameters which will make network suffer from overfitting. The two main advantages of SPPN compared with prediction network in CHAR\cite{bahdanau2014neural} are integrating global semantic information and avoiding huge number of parameters. 
We adopt SPPN as the prediction network for CSTR  with overall consideration.

\begin{table*}[ht]
    \centering
    \caption{Comparisons of scene text recognition performance with previous methods on several benchmarks. $^{\dagger}$ means the method evaluate with beam search, $^{\natural}$ means the method evaluate with TTA. $^{*}$ means the corresponding measured dataset is unclear.}
    \label{table.SOTA}
    \resizebox{\textwidth}{!}{
	\begin{tabular}{ | c | c | c | c | c | c | c | c | c | c | c | c| c | c |}
		\hline
		 	& Method 	& Data	 & Annos & IIIT5K & SVT & \multicolumn{2}{c|}{IC03} & \multicolumn{2}{c|}{IC13} & \multicolumn{2}{c|}{IC15} & \multicolumn{2}{c|}{SVTP} \\ 
		\hline
		    &    &    &    & 3000 & 647 & 860 & 867 & 857 & 1015 & 1811 & 2077 & 639 & 645 \\
		\hline
		\multirow{10}{*}{Seq2seq}
		    & AON \cite{cheng2018aon} & MJ + ST & word & 87.0 & 82.8 & - & 91.5 & - & - & - & 68.2 & 73.0 & -\\ 
		    & FAN \cite{cheng2017focusing} & MJ +ST & word & 87.4 & 85.9& - & 94.2 & - & 93.3 &70.6 & - & - & -\\
		    & Baek \textit{et al.}\cite{baek2019wrong} & MJ+ST & word & 87.9 & 87.5& 94.9 & 94.4 & 93.6 & 92.3 & 77.6 &  71.8 & - & 79.2 \\
		    & SEED \cite{qiao2020seed} $^{\dagger}$ & MJ + ST & word & 93.8 & 89.6 & - & - & - & 92.8 & - & $80^{*}$ & - & 81.4 \\ 
		    & SRN \cite{yu2020towards} & MJ + ST & word & \textbf{94.8} & 91.5 & - & - & - & \textbf{95.5} & 82.7 & - & 85.1 & - \\ 
		    & ASTER\cite{shi2018aster} $^{\dagger}$ & MJ + ST & word & 93.4 & 89.5 & 94.5 & - & - & 91.8 & - & $76.1^{*}$ & 78.5 & -\\ 
		    & SAR\cite{li2019show} $^{\dagger\natural}$ & MJ + ST & word & 91.5 & 84.5 & - & - & - & 91.0 & - & 69.2 & 76.4 & - \\ 
		    &SATRN \cite{lee2020recognizing} & MJ+ST & word & 92.8 & 91.3 & - & \textbf{96.7} & - & 94.1 & - & 79.0 & - & \textbf{86.5}\\
		    & DAN \cite{wang2020decoupled} & MJ+ST & word & 94.3 & 89.2 & - & 95.0 & - & 93.9 & - & 74.5 & 80.0 & - \\
		\hline
		\multirow{3}{*}{Segmentation}
		    & CA-FCN \cite{liao2019scene} & ST & word,char & 91.9 & 86.4 & - & - & - & 91.5 & - & - & - & - \\ 
		    & TextScanner\cite{wan2020textscanner} & MJ + ST & word,char & 93.9 & 90.1 & - & - & - & 92.9 & - & 79.4 & 84.3 & - \\ 
		\hline
		\multirow{3}{*}{Classification}
		    & CHAR\cite{jaderberg14c} & MJ & word & - & 68.0 & - & - & - & 79.5 & - & - & - & - \\ 
		    & CSTR & MJ + ST & word & 93.7 & 90.1 & 95 & 94.8 & 95.3 & 93.2 & 85.6 & 81.6 & - & 85 \\ 
		    & STN-CSTR & MJ + ST & word & 94.2 & \textbf{92.3} & \textbf{95.3} & 95.4 & \textbf{96.3} & 94.1 & \textbf{86.1} & \textbf{82.0} & - & 86.2 \\
		\hline 
	\end{tabular}
	} 
\end{table*}

\section{Experiments}

\subsection{Datasets}

\subsubsection{Training}
We only use two common synthetic text datasets as the training data in our experiments:

\textbf{MJSynth} (MJ) \cite{jaderberg14c} is a synthetic text in image dataset which contains 9 million word box images, generated from a lexicon of 90K English words.

\textbf{SynthText} (ST) \cite{gupta2016synthetic} is a synthetic text in image dataset, designed for scene text detection and recognition. It contains 8 million text boxes from 800K synthetic scene images.
\subsubsection{Testing}

All experiments are evaluated on the six Latin scene text benchmarks described bellow, which contain both regular and irregular text:

\textbf{ICDAR 2003} (IC03) \cite{lucas2005icdar} was created for the ICDAR 2003 Robust Reading competition for reading camera-captured scene texts. It contains 1110 images for evaluation. Following previous work\cite{baek2019wrong, lee2020recognizing}, we ignore all words that are either too short(less than 3 characters) or ones that contain non-alphanumeric characters. Thus, we got 867 images for evaluation.

\textbf{ICDAR 2013} (IC13) \cite{karatzas2013icdar} contains 1095 images for evaluation, where pruning words with non-alphanumeric characters results in 1015 images.

\textbf{ICDAR 2015} (IC15) \cite{karatzas2015icdar} was created for the ICDAR 2015 Robust Reading competitions and contains 2077 images for evaluation. The images are taken with Google Glasses without careful position and focusing.

\textbf{IIIT 5K-Words} (IIIT5k) \cite{mishra2012scene} is collected from the website. It contains 3000 test images for evaluation.

\textbf{Street View Text} (SVT) \cite{wang2011end} has 647 testing images cropped form Google Street View. Many images are severely corrupted by noise, blur, and low resolution.

\textbf{Street View Text-Perspective }(SVTP) \cite{phan2013recognizing} is also cropped form Google Street View. There are 645 test images in this set and many of them are perspectively distorted.

\subsection{Implementation Details}
\label{section.implementation_details}

\textbf{General Setting.}
The size of input images is set to $48\times192$. The number of character classes is set to 37, including 26 alphabets, 10 digitals and 1 end token. And the max length of output sequence $k$ is set to 25.

\textbf{Data Augmentation.}
We adopt some common image processing operations, such as motion blur, gaussian noise and color jitter, and randomly add them to the training images.

\textbf{Model Training.}
The proposed model is trained from scratch without finetuning on other datasets. Label smoothing and warming up are used for all experiments. The batch size is set to 192. Adadelta optimizer is adopted with the initial learning rate 1. The model is totally trained for 420k iterations and the learning rate is decreased $10^{-1}$ and $10^{-2}$ at 150k iterations and 250k iterations. All experiments are implemented on a workstation with 10 NVIDIA 1080Ti graphical cards.

\textbf{Model Testing.}
At runtime, images are rescaled to $48\times192$ without keeping ratio. Unlike \cite{li2019show, litman2020scatter}, we do not use test time augmentation like beam-search or rotating images whose height is larger than width. The predicted text line is calculated by taking the highest probability character at each head and remove the end token.

\subsection{Comparisons with State-of-the-Arts}

The comparisons of our method with previous methods on various scene text recognition benchmarks are shown in Table \ref{table.SOTA}.  We only compare the results without any lexicon, because the lexicon is always unknown before recognition in practical use.

As for classification-based methods, it is evident that our approach outperforms CHAR \cite{jaderberg14c} by a large margin which demonstrates our superiority over CHAR. In particular, our approach gives accuracy increases of $22.1\%$ ($68.0\%$ to $90.1\%$) on SVT and $13.7\%$ gains ($79.5\%$ to $93.2\%$) on IC13.

Compared with seq2seq-based and segmentation-based methods, our approach also achieves comparable performance although 
CSTR is just a simple image classification method without STN \cite{shi2018aster}.
Actually, our model performs the best on 1 of the 6 evaluated text settings which shows the feasibility of perspective from image classification. The advantage of our model is obvious, it is easy to implement and simple for both training and testing phases. 
Furthermore, STN-CSTR performs the best on two benchmarks (SVT, IC15) and the second best on three benchmarks (IC03, IC13, SVTP) compared with previous works.

\subsection{Ablation Study}
\label{section.ablation_study}
We perform a thorough ablation study to present our design concept. For metric evaluation, we provide the average accuracy like \cite{baek2019wrong} does on the unified test dataset involving all subsets of the above mentioned benchmarks. We use the same codebase to run all experiments. We set training iterations to 300k in this section for efficiency.

First, we compare our classification-based method with CTC which is widely used and practical in production \cite{borisyuk2018rosetta,du2020pp}.
Second, we design CPNet to suit the image classification method which is demonstrated to be very important.
Third, we find that data augmentation is crucial to scene text recognition which has not received the attention it deserves.

\subsubsection{From CTC Loss to CE Loss}
\label{section.from_CTC_loss_to_CE_loss}

CTC is a way to get around not knowing the alignment between the input and the output which is an intermediate process. However, \textit{is this intermediate process necessary}?

To answer this question, we conduct the first experiment where we substitute the CTC loss with cross entropy (CE) loss directly and use SHPN as prediction network without other settings changed on the basic backbone network \cite{cheng2017focusing} as mentioned in Section \ref{section.bacbone_network}, the result of which is shown in the first row of Table \ref{table.CTC_vs_CE}. Here we set input resolution to $32\times100$ and no data augmentation is used for fair comparison, other settings are the same as described in Section \ref{section.implementation_details}. The average accuracy of CE loss is almost the same as CTC loss which indicates that \textbf{the intermediate process of alignment between the input and the output is unnecessary}.

The mechanism behind CE loss is that $i$-th classification head predicts $i$-th character of the word sequence in the input image, where the classification head should know the $i$ characters from the left. Intuitively, global receptive field is needed for the classification heads in CE loss based model. We perform the second experiment where the features output by backbone are fed into SPPN which is depicted in the third row of Table \ref{table.CTC_vs_CE}. Considering that there are two gaps between SHPN and SPPN, \textit{i.e.} number of parameters and the presence of average pooling layer, we conduct a third experiment (SEPN) to eliminate the influence of parameters. The above three experiments prove our intuition: \textbf{global receptive field is needed for the prediction network in CE loss based model}.

As shown in Table \ref{table.CTC_vs_CE}, CTC gets the best accuracy $83.8\%$ with SHPN. While for CE, SPPN achives best accuracy $84.1\%$. This clearly reveals that CE can work better than CTC when the prediction network design is proper. 
We use the basic backbone network and SPPN as the base model for all the remaining experiments.

\begin{table}[htb]
	\centering
    \caption{Comparisons of the performance between CTC and CE with different prediction network.}
    \label{table.CTC_vs_CE}
    	\begin{tabular}{ | c | c | c |}
    		\hline
    		 	& Method & Average \\ 
    		\hline
    		\multirow{2}{*}{SHPN}
    		    & CTC & 83.8 \\ 
    		    & CE & 83.6 \\ 
    		\hline
    		\multirow{2}{*}{SEPN}
    		    & CTC & 83.2 \\ 
    		    & CE & 83.2 \\ 
    		\hline 
    		\multirow{2}{*}{SPPN}
    		    & CTC & 82.4 \\ 
    		    & CE & \textbf{84.1} \\ 
    		\hline
    	\end{tabular}
	
\end{table}

\subsubsection{CPNet}
The design of backbone architecture varies according to different perspectives and has direct impact on performance. Since the backbone network in the base model is designed under the seq2seq perspective, which is not suitable for our modeling method, it is necessary for us to redesign the backbone network. 

Inspired by works in image classification and object detection, we consider to adapt the backbone network through model scaling, receptive field, attention mechanism, downsampling accordingly. Since there exist various effective methods and modules \cite{tan2019efficientnet,woo2018cbam,lin2017feature}, we directly use some of them (EM as mentioned in Section \ref{section.bacbone_network}) to scale model capacity, enlarge receptive field and increase complexity of feature interaction.
Moreover, to alleviate the loss of discriminative information in downsampling, we propose the new module named as SADM. 

To evaluate the effectiveness of the two parts in our model, we conduct a series of experiments with/without them. As shown in Table \ref{table.model_redesign}, with general enhanced modules, we achieve $3.1\%$ ($84.1\%$ to $87.2\%$) improvement in average accuracy. Equipped with SADM, we get further boost in performance. Table \ref{table.model_redesign} clearly reveals the importance and effectiveness of our redesigned network CPNet.

\begin{table}[ht]
    \caption{Ablation study of CPNet. ``Base" means the basic CE-based model with SPPN.
    ``EM" means the enhanced module. ``SADM" means the semantic-aware downsampling module. CPNet = Base + EM + SADM.}
    \label{table.model_redesign}
	\centering
	\begin{tabular}{ | c | c |}
		\hline
		 	Method & Average \\ 
		\hline
		Base & 84.1 \\
		\hline
		Base + EM & 87.2 \\
		\hline
		Base + EM + SADM & \textbf{87.3}\\
		\hline
	\end{tabular}
\end{table}

\subsubsection{Data Augmentation}
Data augmentation is widely used in the training of deep CNN. It is an explicit form of regularization and aims to artificially enlarge the training dataset from existing data using data transforms. As is known to all, data augmentation plays an important role in many computer vision tasks, such as image classification, object detection, semantic segmentation and so on. Some researchers are dedicated to proposing specific data augmentation methods. Although there are a significant amount of synthetic data for training in scene text recognition, the influence of data augmentation in it has been ignored in previous works. Therefore, to evaluate how data augmentation influences performance, we conduct the experiment and choose some common operations as mentioned in Section \ref{section.implementation_details}. 

As depicted in Table \ref{table.data_augmentation}, without data augmentation, the average accuracy drops $1.7\%$, which demonstrates that data augmentation is crucial to scene text recognition like other tasks. To the best of our knowledge, we are the first one to reveal the importance of data augmentation for scene text recognition. We hope this finding may inspire researchers to do further explorations on data augmentation in this field.

\begin{table}[ht]
    \caption{Ablation Study of data augmentation.  ``Base" means CE loss with CPNet and SPPN. ``DA" means data augmentation.}
    \label{table.data_augmentation}
	\centering
	\begin{tabular}{ | c | c |}
		\hline
		 	Method & Average \\ 
		\hline
		Base & 87.3\\
		\hline
		Base + DA & \textbf{89.0}\\
		\hline
	\end{tabular}
\end{table}

\subsubsection{Spatial Transformer Network}
When a spatial transformer network is added upon our CSTR to rectify the input image, which is called STN-CSTR, $0.4\%$ perfomance boost can be achieved, as is showed in Table \ref{table.stn}.

\begin{table}[ht]
    \caption{Ablation Study of STN.}
    \label{table.stn}
	\centering
	\begin{tabular}{ | c | c |}
		\hline
		 	Method & Average \\ 
		\hline
		CSTR & 89.0\\
		\hline
		CSTR + STN & \textbf{89.4}\\
		\hline
	\end{tabular}
\end{table}

\section{Conclusion}
In this paper, we revisit classification perspective on scene text recognition, which models the scene text recognition as an image classification problem. Based on the image classification perspective, we design a scene text recognition model, which is named as CSTR.
The CSTR model consists of 
CPNet and SPPN which is composed of a global average pooling layer and independent convolutional layers, each of which predicts the corresponding character in the text line.
CSTR is as simple as image classification models such as ResNet \cite{he2016deep} which makes it easy to implement, and the fully convolutional neural network architecture makes it easy to train and deploy. 
Furthermore, CSTR achieves comparable performance and STN-CSTR achieves the state-of-the-art performance.
In the future, we are interested in further improving the performance of classification-based methods.

{\small
\bibliographystyle{ieee_fullname}
\bibliography{CSTR}

\begin{thebibliography}{10}\itemsep=-1pt

\bibitem{baek2019wrong}
Jeonghun Baek, Geewook Kim, Junyeop Lee, Sungrae Park, Dongyoon Han, Sangdoo
  Yun, Seong~Joon Oh, and Hwalsuk Lee.
\newblock What is wrong with scene text recognition model comparisons? dataset
  and model analysis.
\newblock In {\em Proceedings of the IEEE/CVF International Conference on
  Computer Vision}, pages 4715--4723, 2019.

\bibitem{bahdanau2014neural}
Dzmitry Bahdanau, Kyunghyun Cho, and Yoshua Bengio.
\newblock Neural machine translation by jointly learning to align and
  translate.
\newblock {\em arXiv preprint arXiv:1409.0473}, 2014.

\bibitem{bartz2019kiss}
Christian Bartz, Joseph Bethge, Haojin Yang, and Christoph Meinel.
\newblock Kiss: Keeping it simple for scene text recognition.
\newblock {\em arXiv preprint arXiv:1911.08400}, 2019.

\bibitem{borisyuk2018rosetta}
Fedor Borisyuk, Albert Gordo, and Viswanath Sivakumar.
\newblock Rosetta: Large scale system for text detection and recognition in
  images.
\newblock In {\em Proceedings of the 24th ACM SIGKDD International Conference
  on Knowledge Discovery \& Data Mining}, pages 71--79, 2018.

\bibitem{brown2020language}
Tom~B Brown, Benjamin Mann, Nick Ryder, Melanie Subbiah, Jared Kaplan, Prafulla
  Dhariwal, Arvind Neelakantan, Pranav Shyam, Girish Sastry, Amanda Askell,
  et~al.
\newblock Language models are few-shot learners.
\newblock {\em arXiv preprint arXiv:2005.14165}, 2020.

\bibitem{carion2020end}
Nicolas Carion, Francisco Massa, Gabriel Synnaeve, Nicolas Usunier, Alexander
  Kirillov, and Sergey Zagoruyko.
\newblock End-to-end object detection with transformers.
\newblock {\em arXiv preprint arXiv:2005.12872}, 2020.

\bibitem{cheng2017focusing}
Zhanzhan Cheng, Fan Bai, Yunlu Xu, Gang Zheng, Shiliang Pu, and Shuigeng Zhou.
\newblock Focusing attention: Towards accurate text recognition in natural
  images.
\newblock In {\em Proceedings of the IEEE international conference on computer
  vision}, pages 5076--5084, 2017.

\bibitem{cheng2018aon}
Zhanzhan Cheng, Yangliu Xu, Fan Bai, Yi Niu, Shiliang Pu, and Shuigeng Zhou.
\newblock Aon: Towards arbitrarily-oriented text recognition.
\newblock In {\em Proceedings of the IEEE Conference on Computer Vision and
  Pattern Recognition}, pages 5571--5579, 2018.

\bibitem{chorowski2015attention}
Jan~K Chorowski, Dzmitry Bahdanau, Dmitriy Serdyuk, Kyunghyun Cho, and Yoshua
  Bengio.
\newblock Attention-based models for speech recognition.
\newblock {\em Advances in neural information processing systems}, 28:577--585,
  2015.

\bibitem{devlin2018bert}
Jacob Devlin, Ming-Wei Chang, Kenton Lee, and Kristina Toutanova.
\newblock Bert: Pre-training of deep bidirectional transformers for language
  understanding.
\newblock {\em arXiv preprint arXiv:1810.04805}, 2018.

\bibitem{du2020pp}
Yuning Du, Chenxia Li, Ruoyu Guo, Xiaoting Yin, Weiwei Liu, Jun Zhou, Yifan
  Bai, Zilin Yu, Yehua Yang, Qingqing Dang, et~al.
\newblock Pp-ocr: A practical ultra lightweight ocr system.
\newblock {\em arXiv preprint arXiv:2009.09941}, 2020.

\bibitem{gao2018recurrent}
Yunze Gao, Yingying Chen, Jinqiao Wang, Zhen Lei, Xiao-Yu Zhang, and Hanqing
  Lu.
\newblock Recurrent calibration network for irregular text recognition.
\newblock {\em arXiv preprint arXiv:1812.07145}, 2018.

\bibitem{graves2006connectionist}
Alex Graves, Santiago Fern{\'a}ndez, Faustino Gomez, and J{\"u}rgen
  Schmidhuber.
\newblock Connectionist temporal classification: labelling unsegmented sequence
  data with recurrent neural networks.
\newblock In {\em Proceedings of the 23rd international conference on Machine
  learning}, pages 369--376, 2006.

\bibitem{gupta2016synthetic}
Ankush Gupta, Andrea Vedaldi, and Andrew Zisserman.
\newblock Synthetic data for text localisation in natural images.
\newblock In {\em Proceedings of the IEEE conference on computer vision and
  pattern recognition}, pages 2315--2324, 2016.

\bibitem{he2016deep}
Kaiming He, Xiangyu Zhang, Shaoqing Ren, and Jian Sun.
\newblock Deep residual learning for image recognition.
\newblock In {\em Proceedings of the IEEE conference on computer vision and
  pattern recognition}, pages 770--778, 2016.

\bibitem{hu2020gtc}
Wenyang Hu, Xiaocong Cai, Jun Hou, Shuai Yi, and Zhiping Lin.
\newblock Gtc: Guided training of ctc towards efficient and accurate scene text
  recognition.
\newblock In {\em AAAI}, pages 11005--11012, 2020.

\bibitem{jaderberg14c}
Max Jaderberg, Karen Simonyan, Andrea Vedaldi, and Andrew Zisserman.
\newblock Synthetic data and artificial neural networks for natural scene text
  recognition.
\newblock In {\em Workshop on Deep Learning, NIPS}, 2014.

\bibitem{jaderberg2015spatial}
Max Jaderberg, Karen Simonyan, Andrew Zisserman, et~al.
\newblock Spatial transformer networks.
\newblock {\em Advances in neural information processing systems},
  28:2017--2025, 2015.

\bibitem{karatzas2015icdar}
Dimosthenis Karatzas, Lluis Gomez-Bigorda, Anguelos Nicolaou, Suman Ghosh,
  Andrew Bagdanov, Masakazu Iwamura, Jiri Matas, Lukas Neumann,
  Vijay~Ramaseshan Chandrasekhar, Shijian Lu, et~al.
\newblock Icdar 2015 competition on robust reading.
\newblock In {\em 2015 13th International Conference on Document Analysis and
  Recognition (ICDAR)}, pages 1156--1160. IEEE, 2015.

\bibitem{karatzas2013icdar}
Dimosthenis Karatzas, Faisal Shafait, Seiichi Uchida, Masakazu Iwamura,
  Lluis~Gomez i Bigorda, Sergi~Robles Mestre, Joan Mas, David~Fernandez Mota,
  Jon~Almazan Almazan, and Lluis~Pere De~Las~Heras.
\newblock Icdar 2013 robust reading competition.
\newblock In {\em 2013 12th International Conference on Document Analysis and
  Recognition}, pages 1484--1493. IEEE, 2013.

\bibitem{lecun1998gradient}
Yann LeCun, L{\'e}on Bottou, Yoshua Bengio, and Patrick Haffner.
\newblock Gradient-based learning applied to document recognition.
\newblock {\em Proceedings of the IEEE}, 86(11):2278--2324, 1998.

\bibitem{lee2020recognizing}
Junyeop Lee, Sungrae Park, Jeonghun Baek, Seong Joon~Oh, Seonghyeon Kim, and
  Hwalsuk Lee.
\newblock On recognizing texts of arbitrary shapes with 2d self-attention.
\newblock In {\em Proceedings of the IEEE/CVF Conference on Computer Vision and
  Pattern Recognition Workshops}, pages 546--547, 2020.

\bibitem{li2019show}
Hui Li, Peng Wang, Chunhua Shen, and Guyu Zhang.
\newblock Show, attend and read: A simple and strong baseline for irregular
  text recognition.
\newblock In {\em Proceedings of the AAAI Conference on Artificial
  Intelligence}, volume~33, pages 8610--8617, 2019.

\bibitem{liao2019scene}
Minghui Liao, Jian Zhang, Zhaoyi Wan, Fengming Xie, Jiajun Liang, Pengyuan Lyu,
  Cong Yao, and Xiang Bai.
\newblock Scene text recognition from two-dimensional perspective.
\newblock In {\em Proceedings of the AAAI Conference on Artificial
  Intelligence}, volume~33, pages 8714--8721, 2019.

\bibitem{lin2017feature}
Tsung-Yi Lin, Piotr Doll{\'a}r, Ross Girshick, Kaiming He, Bharath Hariharan,
  and Serge Belongie.
\newblock Feature pyramid networks for object detection.
\newblock In {\em Proceedings of the IEEE conference on computer vision and
  pattern recognition}, pages 2117--2125, 2017.

\bibitem{lin2017structured}
Zhouhan Lin, Minwei Feng, Cicero Nogueira~dos Santos, Mo Yu, Bing Xiang, Bowen
  Zhou, and Yoshua Bengio.
\newblock A structured self-attentive sentence embedding.
\newblock {\em arXiv preprint arXiv:1703.03130}, 2017.

\bibitem{litman2020scatter}
Ron Litman, Oron Anschel, Shahar Tsiper, Roee Litman, Shai Mazor, and R
  Manmatha.
\newblock Scatter: selective context attentional scene text recognizer.
\newblock In {\em Proceedings of the IEEE/CVF Conference on Computer Vision and
  Pattern Recognition}, pages 11962--11972, 2020.

\bibitem{liu2016star}
Wei Liu, Chaofeng Chen, Kwan-Yee~K Wong, Zhizhong Su, and Junyu Han.
\newblock Star-net: A spatial attention residue network for scene text
  recognition.
\newblock In {\em BMVC}, volume~2, page~7, 2016.

\bibitem{lu2019master}
Ning Lu, Wenwen Yu, Xianbiao Qi, Yihao Chen, Ping Gong, and Rong Xiao.
\newblock Master: Multi-aspect non-local network for scene text recognition.
\newblock {\em arXiv preprint arXiv:1910.02562}, 2019.

\bibitem{lucas2005icdar}
Simon~M Lucas, Alex Panaretos, Luis Sosa, Anthony Tang, Shirley Wong, Robert
  Young, Kazuki Ashida, Hiroki Nagai, Masayuki Okamoto, Hiroaki Yamamoto,
  et~al.
\newblock Icdar 2003 robust reading competitions: entries, results, and future
  directions.
\newblock {\em International Journal of Document Analysis and Recognition
  (IJDAR)}, 7(2-3):105--122, 2005.

\bibitem{mishra2012scene}
Anand Mishra, Karteek Alahari, and CV Jawahar.
\newblock Scene text recognition using higher order language priors.
\newblock In {\em BMVC-British Machine Vision Conference}. BMVA, 2012.

\bibitem{phan2013recognizing}
Trung~Quy Phan, Palaiahnakote Shivakumara, Shangxuan Tian, and Chew~Lim Tan.
\newblock Recognizing text with perspective distortion in natural scenes.
\newblock In {\em Proceedings of the IEEE International Conference on Computer
  Vision}, pages 569--576, 2013.

\bibitem{qiao2020seed}
Zhi Qiao, Yu Zhou, Dongbao Yang, Yucan Zhou, and Weiping Wang.
\newblock Seed: Semantics enhanced encoder-decoder framework for scene text
  recognition.
\newblock In {\em Proceedings of the IEEE/CVF Conference on Computer Vision and
  Pattern Recognition}, pages 13528--13537, 2020.

\bibitem{sheng2019nrtr}
Fenfen Sheng, Zhineng Chen, and Bo Xu.
\newblock Nrtr: A no-recurrence sequence-to-sequence model for scene text
  recognition.
\newblock In {\em 2019 International Conference on Document Analysis and
  Recognition (ICDAR)}, pages 781--786. IEEE, 2019.

\bibitem{shi2016end}
Baoguang Shi, Xiang Bai, and Cong Yao.
\newblock An end-to-end trainable neural network for image-based sequence
  recognition and its application to scene text recognition.
\newblock {\em IEEE transactions on pattern analysis and machine intelligence},
  39(11):2298--2304, 2016.

\bibitem{shi2016robust}
Baoguang Shi, Xinggang Wang, Pengyuan Lyu, Cong Yao, and Xiang Bai.
\newblock Robust scene text recognition with automatic rectification.
\newblock In {\em Proceedings of the IEEE conference on computer vision and
  pattern recognition}, pages 4168--4176, 2016.

\bibitem{shi2018aster}
Baoguang Shi, Mingkun Yang, Xinggang Wang, Pengyuan Lyu, Cong Yao, and Xiang
  Bai.
\newblock Aster: An attentional scene text recognizer with flexible
  rectification.
\newblock {\em IEEE transactions on pattern analysis and machine intelligence},
  41(9):2035--2048, 2018.

\bibitem{tan2019efficientnet}
Mingxing Tan and Quoc Le.
\newblock Efficientnet: Rethinking model scaling for convolutional neural
  networks.
\newblock In {\em International Conference on Machine Learning}, pages
  6105--6114. PMLR, 2019.

\bibitem{vaswani2017attention}
Ashish Vaswani, Noam Shazeer, Niki Parmar, Jakob Uszkoreit, Llion Jones,
  Aidan~N Gomez, {\L}ukasz Kaiser, and Illia Polosukhin.
\newblock Attention is all you need.
\newblock In {\em Advances in neural information processing systems}, pages
  5998--6008, 2017.

\bibitem{wan2020textscanner}
Zhaoyi Wan, Minghang He, Haoran Chen, Xiang Bai, and Cong Yao.
\newblock Textscanner: Reading characters in order for robust scene text
  recognition.
\newblock In {\em Proceedings of the AAAI Conference on Artificial
  Intelligence}, pages 12120--12127, 2020.

\bibitem{wang2011end}
Kai Wang, Boris Babenko, and Serge Belongie.
\newblock End-to-end scene text recognition.
\newblock In {\em 2011 International Conference on Computer Vision}, pages
  1457--1464. IEEE, 2011.

\bibitem{wang2020decoupled}
Tianwei Wang, Yuanzhi Zhu, Lianwen Jin, Canjie Luo, Xiaoxue Chen, Yaqiang Wu,
  Qianying Wang, and Mingxiang Cai.
\newblock Decoupled attention network for text recognition.
\newblock In {\em Proceedings of the AAAI Conference on Artificial
  Intelligence}, pages 12216--12224, 2020.

\bibitem{wang2018non}
Xiaolong Wang, Ross Girshick, Abhinav Gupta, and Kaiming He.
\newblock Non-local neural networks.
\newblock In {\em Proceedings of the IEEE conference on computer vision and
  pattern recognition}, pages 7794--7803, 2018.

\bibitem{woo2018cbam}
Sanghyun Woo, Jongchan Park, Joon-Young Lee, and In~So Kweon.
\newblock Cbam: Convolutional block attention module.
\newblock In {\em Proceedings of the European conference on computer vision
  (ECCV)}, pages 3--19, 2018.

\bibitem{yang2019simple}
Lu Yang, Peng Wang, Hui Li, Ye Gao, Linjiang Zhang, Chunhua Shen, and Yanning
  Zhang.
\newblock A simple and strong convolutional-attention network for irregular
  text recognition.
\newblock {\em arXiv preprint arXiv:1904.01375}, 2019.

\bibitem{yang2019symmetry}
Mingkun Yang, Yushuo Guan, Minghui Liao, Xin He, Kaigui Bian, Song Bai, Cong
  Yao, and Xiang Bai.
\newblock Symmetry-constrained rectification network for scene text
  recognition.
\newblock In {\em Proceedings of the IEEE International Conference on Computer
  Vision}, pages 9147--9156, 2019.

\bibitem{yu2020towards}
Deli Yu, Xuan Li, Chengquan Zhang, Tao Liu, Junyu Han, Jingtuo Liu, and Errui
  Ding.
\newblock Towards accurate scene text recognition with semantic reasoning
  networks.
\newblock In {\em Proceedings of the IEEE/CVF Conference on Computer Vision and
  Pattern Recognition}, pages 12113--12122, 2020.

\bibitem{zhan2019esir}
Fangneng Zhan and Shijian Lu.
\newblock Esir: End-to-end scene text recognition via iterative image
  rectification.
\newblock In {\em Proceedings of the IEEE Conference on Computer Vision and
  Pattern Recognition}, pages 2059--2068, 2019.

\end{thebibliography}
}

\end{document}